%% file: main.tex
\documentclass{article}
\usepackage{spconf,amsmath,graphicx}
\usepackage{amssymb}
\usepackage{caption}
\usepackage{subcaption}
\usepackage{multirow}
\usepackage{bm}
\makeatletter
\let\NAT@parse\undefined
\makeatother
\usepackage[colorlinks,linkcolor=red,anchorcolor=blue,urlcolor=magenta,citecolor=blue]{hyperref}

\title{Elevating Skeleton-Based Action Recognition with Efficient Multi-Modality Self-Supervision}
\name{
\begin{tabular}{@{}c@{}}
Yiping Wei$^{1}$ \qquad 
Kunyu Peng$^{1,*}$ \qquad 
Alina Roitberg$^{2}$ \qquad
Jiaming Zhang$^{1}$ \qquad
Junwei Zheng$^{1}$ \\
Ruiping Liu$^{1}$ \qquad
Yufan Chen$^{1}$ \qquad
Kailun Yang$^{3}$ \qquad
Rainer Stiefelhagen$^{1}$
\end{tabular}
\address{$^{1}$Institute for Anthropomatics and Robotics, Karlsruhe Institute of Technology, Germany.\\$^{2}$Institute for Artificial Intelligence, University of Stuttgart, Germany.\\$^{3}$School of Robotics, Hunan University, China.}
\thanks{The project served to prepare the SFB 1574 Circular Factory for the Perpetual Product (project ID: 471687386), approved by the Deutsche Forschungsgemeinschaft (DFG, German Research Foundation) with a start date of April 1, 2024. This work was also partially supported in part by the SmartAge project sponsored by the Carl Zeiss Stiftung (P2019-01-003; 2021-2026).}
\thanks{*Corresponding Author (Email: kunyu.peng@kit.edu).}
}

\begin{document}
%\ninept
%
\maketitle
\begin{abstract}
Self-supervised representation learning for human action recognition has developed rapidly in recent years. Most of the existing works are based on skeleton data while using a multi-modality setup. These works overlooked the differences in performance among modalities, which led to the propagation of erroneous knowledge between modalities while only three fundamental modalities, \textit{i.e.}, \emph{joints}, \emph{bones}, and \emph{motions} are used, hence no additional modalities are explored.

In this work, we first propose an \textbf{I}mplicit \textbf{K}nowledge \textbf{E}xchange \textbf{M}odule (IKEM) which alleviates the propagation of erroneous knowledge between low-performance modalities. Then, we further propose three new modalities to enrich the complementary information between modalities. Finally, to maintain efficiency when introducing new modalities, we propose a novel teacher-student framework to distill the knowledge from the secondary modalities into the mandatory modalities considering the relationship constrained by anchors, positives, and negatives, named relational cross-modality knowledge distillation. The experimental results demonstrate the effectiveness of our approach, unlocking the efficient use of skeleton-based multi-modality data. Source code will be made publicly available at \url{https://github.com/desehuileng0o0/IKEM}.
\end{abstract}
\section{Introduction}
\label{sec:intro}

The task of self-supervised human action recognition represents a critical endeavor within the domain of computer vision~\cite{li20213d}.
It centers on the understanding of human behavior and the assignment of labels to observed actions without the dependency on labeled data during the training process. This approach not only carries profound research implications but also harbors significant research promise across a multitude of domains, including video surveillance \cite{ziaeefard2015semantic}, human-computer interaction \cite{presti20163d}, and virtual reality \cite{fangbemi2018efficient}. Additionally, it stands out for its ability to alleviate the burden of human annotation considering the large-scale dataset.
Existing self-supervised action recognition approaches can be clustered into video- and skeleton-based approaches, where video-based approaches~\cite{xu2019self,dave2022spact} are computationally expensive and not privacy-preserving.
Considering the aforementioned issue, self-supervised skeleton-based action recognition~\cite{guo2021contrastive,zhou2023selfsupervised,lin2023actionlet,zhang2022contrastive} plays an important role when the RGB data is sensible. 

Unlike the works for the supervised skeleton-based human action recognition which focuses mostly on developing superior model architectures~\cite{duan2022pyskl,liu2023temporal,trivedi2022psumnet, Huang2023GraphCL, li2019actionalstructural, duan2022revisiting}, the works for self-supervised skeleton-based human action recognition focus more on the self-training strategies. A variety of contrastive learning frameworks~\cite{chen2020simple,grill2020bootstrap,he2020momentum,oord2019representation,caron2021unsupervised,chen2020improved,chen2020exploring} have been proposed and several applications have appeared with success in the area of self-supervised skeleton-based action recognition. 
CrosSCLR~\cite{li20213d} proposed cross-modality knowledge exchange for the first time. 
Subsequent contrastive learning frameworks for self-supervised skeleton-based action representation learning~\cite{guo2021contrastive,zhou2023selfsupervised,lin2023actionlet,zhang2022contrastive} are proposed to adapt three fundamental modalities of CrosSCLR~\cite{li20213d}. 
However, none of these works adapted the cross-modality knowledge exchange mechanism introduced by CrosSCLR~\cite{li20213d}.
This mechanism leverages additional positive samples with high confidence to guide contrastive learning, explicitly exchanging knowledge between modalities but overlooks the performance gap among them. In some cases, consistent knowledge across modalities boosts confidence, while divergent knowledge relies solely on one modality, potentially leading to erroneous cross-modality knowledge transfer. This performance gap can hinder effective knowledge propagation.
Apart from the knowledge exchange mechanism, all of these existing works rely on three fundamental modalities derived from the original skeleton data, \textit{i.e.}, joints, bones, and motions. However, it is still an open challenge regarding how to use more modalities while considering the efficient usage of the multi-modality data. 

First, to tackle the issue of the explicit knowledge exchange, we propose the \textbf{I}mplicit \textbf{K}nowledge \textbf{E}xchange \textbf{M}odule (IKEM). IKEM does not mine additional positive samples but evaluates all modalities together when computing similarity between samples, transferring knowledge implicitly. During knowledge exchange we encourage each modality to explore knowledge from other modalities at the same time, achieving the transfer of implicit knowledge. Extensive experiments show that implicit knowledge exchange can work simultaneously with explicit knowledge exchange, resulting in a significant improvement in the quality of cross-modality knowledge exchange. Then, we delve deeper into finding more new modalities that could contribute to the self-supervised skeleton-based human action recognition pipeline. With the improved cross-modality knowledge exchange module, three new modalities have been proposed, \textit{i.e.}, accelerations, rotation axis directions, and joint angular velocities. 
Due to the increase of the model size by using more modalities in an ensemble way, we make use of a teacher-student model to distill the useful information from the branches for secondary modalities into the branches for the mandatory modalities by using negative and positive pairs, which can enable the knowledge transfer beyond the cross-modality anchor representations, while considering the geometric structure in the latent space in terms of the relationship among the anchor sample, negative sample, and positive sample.

By using IKEM together with the proposed teacher-student model, our method produces accuracies of $80.7\%$ and $86.0\%$ in cross-subject and cross-view evaluations. 

\section{Methodology}
\label{ch:method}
\subsection{Implicit Knowledge Exchange Module}
\label{sec:IKEM}

\begin{figure*}[htb]
\centering
\includegraphics[width=17cm]{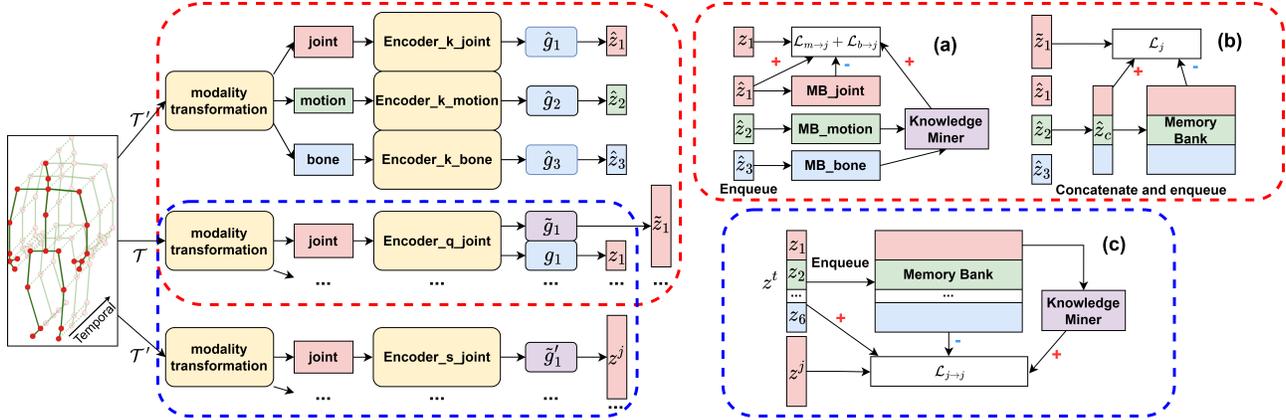}
\vskip-1ex
\caption{An overview of our pre-training model in red dashed box and teacher-student model in blue dashed box, where module (a) is the knowledge exchange module in CrosSCLR, module (b) is our proposed IKEM, and module (c) is the knowledge distillation module for our teacher-student model. $\widetilde{g}$ are the new MLPs introduced by IKEM and $\widetilde{g}'$ denote the MLPs of the student model. MB is the abbreviation for memory bank. All the modules in the figure use the update of the encoder from joint modality as an example.}
\vskip -3ex
\label{fig:3views a and b}
\end{figure*}

We add an additional multi-layer perceptron (MLP) after each query encoder, denoted as $\widetilde{g}_u$, where $u \in \{1,2,...\}$ is the index of different modality.
The intermediate features are mapped into a new latent space formulated as $\widetilde{z}_u=\widetilde{g}_u(f_u) \in \mathbb{R}^{n\times c_z}$, where $n$, $c_z$ and $f_u$ are the number of modalities, the output dimension of the original MLP in purple in Figure~\ref{fig:3views a and b} and the input feature from different modalities, respectively. We leverage $\widetilde{z}_u$ as an anchor point to construct a new contrastive loss~\cite{1640964}. The embedding of the positive samples is a concatenation of the embeddings from the three key encoders, $\hat{z}_c = [\hat{z}_1,\hat{z}_2,\hat{z}_3]$, where $[\cdot,\cdot]$ denotes concatenation operation. At the end of each iteration, the embeddings of this batch will be deposited into a queue-based memory bank $M_c$, while the oldest batch will be discharged. The samples in the memory bank act as negative samples in the computation of the contrastive loss~\cite{1640964}.

In summary, for any modality $u$, we express the loss function formally as
\begin{equation}
\mathcal{L}_u=-\log\frac{\exp(\tilde{z}_u\cdot\hat{z}_c/\tau)}{\exp(\tilde{z}_u\cdot\hat{z}_c/\tau)+\sum_{i\in N}\exp(\tilde{z}_u\cdot \hat{m}_i/\tau)},
\end{equation}
where $\hat{m}_i \in M_c$, $\tau$ is a temperature hyper-parameter. Finally, IKEM has the following objective:
\begin{equation}
\label{eq:lossike}
\mathcal{L}_{\mathrm{ikem}}=\sum_u^U\mathcal{L}_u.
\end{equation}
The cross-modality knowledge exchange of CrosSCLR~\cite{li20213d} directly specifies additional positive samples for modal training, denoted as \textbf{E}xplicit \textbf{K}nowledge \textbf{E}xchange \textbf{M}oudle (EKEM). In contrast, we do not mine any additional positive samples in IKEM but make one modality learn to output the embedding of all modalities. Accordingly, IKEM enables the model to learn mutual information contained in the inputs of different modalities. We consider the additional knowledge transferred as implicit knowledge. Thereby, the implicit cues excavated by our IKEM lie in two parts: (a) the implicit exchange of knowledge and (b) the exchange of implicit knowledge.

\subsection{Three New Modalities} 
We propose three new modalities: acceleration, rotation axis, and angular velocity, along with the three existing modalities to form our six-modality model.
In data pre-processing of the skeleton sequences, we resize all sequences to $50$ frames by linear interpolation. The 3D human skeleton sequence $x\in\mathbb{R}^{C\times T\times V}$ has $T$ frames with $V$ joints, and each joint has $C=3$ coordinate features. For a single sequence, we consider a parameter $\gamma=T / 50$ termed time scale for the sake of time consistency. 
The motion is represented as the temporal displacement between frames: $\mathbf{m}_{\{:,t+1,:\}} = x_{\{:,t+1,:\}}-x_{\{:,t,:\}}$. With time scale $\gamma$, we can represent velocity as $(x_{\{:,t+1,:\}}-x_{\{:,t,:\}})/\gamma$. Similarly, we calculate the acceleration as:
\begin{equation}
\mathbf{a}_{\{:,t,:\}}=\frac{\mathbf{m}_{\{:,t+1,:\}}/\gamma-\mathbf{m}_{\{:,t,:\}}/\gamma}{\gamma}.
\end{equation}
A bone is the distance between two neighboring joints in the same frame: $\mathbf{b}_{\{:,:,v_2\}} = x_{\{:,:,v_2\}}-x_{\{:,:,v_1\}}$.
We define $25$ single-degree-of-freedom hinge joints and their rotation axes. With $\mathbf{b}_{\{:,:,i\}}$ and $\mathbf{b}_{\{:,:,j\}}$ as the bones that make up this joint, the unit direction vector of a rotation axis $\mathbf{r}$ is calculated as:
\begin{equation}
    \mathbf{r}_{\{:,:,i\}}=\frac{\mathbf{b}_{\{:,:,i\}} \times \mathbf{b}_{\{:,:,j\}}}{\|\mathbf{b}_{\{:,:,i\}}\times \mathbf{b}_{\{:,:,j\}}\|}.
\end{equation}
For the angular velocity of these hinge joints $\theta$, firstly, we calculate the joint angle of a particular hinge joint as:
\begin{equation}
    \mathbf{\theta}_{\{:,i\}}=\cos^{-1}\frac{\mathbf{b}_{\{:,:,i\}}\cdot \mathbf{b}_{\{:,:,j\}}}{\|\mathbf{b}_{\{:,:,i\}}\|\cdot\|\mathbf{b}_{\{:,:,j\}}\|}.
\end{equation}
Then the joint angular velocity $\omega$ of a skeleton sequence can be expressed as:
\begin{equation}
    \bm{\omega}_{\{:,t,:\}}=\mathbf{r}_{\{:,t,:\}}\frac{\theta_{\{t+1,:\}}-\theta_{\{t,:\}}}{\gamma}.
\end{equation}

\subsection{Relational Cross-Modality Knowledge Distillation}
\label{subsec:TSmodel}
We propose a cross-modality teacher-student model to ensure the efficiency of multi-modality representation learning by distilling knowledge from our six-modality model to a model trained with only three fundamental modalities. 
Our knowledge distillation framework is based on MoCo~\cite{he2020momentum}.
Compared with the original framework of MoCo, we stop the momentum update of the teacher encoder. This even leads to a large, fully consistent memory bank that is more conducive to the convergence of the student model. With the same contrastive objective function as MoCo, the student embedding $s$ will be drawn closer to the teacher embedding $t$ and pushed further away from the embedding of the negative samples from the memory bank.

With this idea, our student-teacher model is shown in the blue dashed boxes of Figure~\ref{fig:3views a and b}.
The teacher network is our six-modality model and we distill the knowledge from the model trained with six modalities into the model trained with three commonly used modalities, where we leverage CrosSCLR as the baseline. 
We give the objective function for transferring knowledge from modality $v$ of the teacher model to modality $u$ of the student model:
\begin{equation}
\mathcal{L}_{v\to u}=-\log\frac{\exp(z^t\cdot z^u/\tau)+\exp(s_j^us_j^v/\tau)}{\exp(z^t\cdot z^u/\tau)+\sum_{i\in N}\exp(s_i^us_i^v/\tau)},
\end{equation}
where $j$ represents the index of the most similar negative sample mined from modality $v$, and $z^t$ is the concatenation of the embedding of all modalities of the teacher model.
$s_i^v$ and $s_i^u$ belong to the similarity sets $S_v$ and $S_u$ respectively.
Finally, our teacher-student model has the objective function:
\begin{equation}
\mathcal{L}_{\mathrm{ts}}=\sum_u^U\sum_v^V\mathcal{L}_{v\to u}.
\end{equation}
where $U$ and $V$ are the sets of models trained with different modalities of the student and teacher networks, respectively.

\section{Experiments and Results}
\label{sec:Experiments and Results}
\begin{table*}[ht]
\centering
	\begin{center}
        \caption{%
        Experimental results on the NTU-60 dataset~\cite{shahroudy2016ntu}. ``3s'' and ``6s'' represent three-stream fusion and six-stream fusion. Results are expressed in the linear classification accuracy (\%).}
        \vspace{-8pt}
		\input{tables}
	\end{center}
	\vspace{-20pt}
	\label{tab_1}
\end{table*}
\subsection{Dataset}

The NTU-RGB+D 60 (NTU-60) dataset~\cite{shahroudy2016ntu} is a large-scale dataset for human action recognition. 
The samples are categorized into $60$ diverse action categories that encompass a spectrum of actions ranging from commonplace daily activities to interactive behaviors and actions relevant to health.
Two evaluation protocols are recommended, the cross-subject evaluation and cross-view evaluation, considering the variability introduced by subjects and camera viewpoints.

\subsection{Experimental Setting}
For data preprocessing, we resize each skeleton sequence to $50$ frames using linear interpolation, while saving the original number of frames for each sequence for the calculation of additional modalities. SGD with momentum ($0.9$) and weight decay ($0.0001$) is used for optimization.
The mini-batch size is set to $128$.
For data augmentation $\mathcal{T}$, we follow the strategy 
in CrosSCLR~\cite{li20213d} as well as their amplitude coefficients. 

\noindent\textbf{Self-supervised pre-training.} For representation learning, the model is trained for $300$ epochs with a learning rate of $0.1$ (multiplied by $0.1$ at epoch $250$). Pre-training is divided into two stages. Firstly, the InfoNCE loss~\cite{oord2019representation} is used to train each modality individually. In the second stage, IKEM and EKEM are applied to the model together. Each of the two stages is trained for $150$ epochs.
The temperature hyper-parameter $\tau$ of all objective functions is set to $0.07$.
For better efficiency, we leverage the teacher-student model to train $150$ epochs, distilling the knowledge of our six-modality model to the common three-modality model.

\noindent\textbf{Linear Evaluation Protocol.} The models are validated by linear evaluation for action recognition tasks. Query encoders are frozen and attached to a linear classifier, which was trained in a supervised manner for $100$ epochs with a learning rate of $0.1$ (multiplied by $0.1$ at epoch $80$).

\subsection{Results}
\noindent\textbf{Effectiveness of knowledge exchange modules.}
Table~\ref{tab: IKEMworkwithEKEM} showcases the comparison of EKEM and IKEM, alongside their collaborative performance under the cross-view protocol NTU-60~\cite{shahroudy2016ntu}. Our IKEM offers supplementary information to the inherent knowledge exchange module of EKEM.
The combination of IKEM and EKEM achieves a performance enhancement of $0.7\%$ over the baseline CrosSCLR.
The quantitative results under both linear evaluation protocols on NTU-60~\cite{shahroudy2016ntu} are introduced in Table~\ref{tab: our3sand6s}.
In the case of fusing three modalities, $0.2\%$ and $0.7\%$ improvements are obtained under the cross-subject and cross-view evaluation protocols.

\noindent\textbf{Effectiveness of multi-modality fusion.}
The model is ensembled to incorporate six modalities.
We adopt the strategy leveraged in CrosSCLR, wherein the key encoder is frozen for the modalities exhibiting high performance. The loss of the high-performance modalities converges faster. Among the six available modalities, \textit{joints}, \textit{bones}, and \textit{roation axes} are high-performance modalities. 
As shown in Table~\ref{tab: our3sand6s}, our approach demonstrates enhanced efficacy in extracting representations from the six modalities in a non-redundant manner, yielding performance improvements of $3.2\%$ and $2.8\%$ compared to the three-modality counterpart. Table~\ref{tab: our3sand6s single modality} showcases the performance of individual modalities, our six-modality model completely outperforms our three-modality model. This reaffirms the value of utilizing the complementary information provided by multi-modality skeleton data.

\noindent\textbf{Effectiveness of student models.} 
Table~\ref{tab: studentandother} demonstrates performances of our three-modality student model built based on CrosSCLR and its subsequent works. For all methods, ST-GCN~\cite{yan2018spatial} is utilized as the backbone. Therefore, the models trained with different methods have almost the same number of parameters when deployed, the number of parameters of our student model is $2.510M$. In both evaluation scenarios, our methods achieve substantial improvements compared to the baseline CrosSCLR~\cite{li20213d} while obtaining better or similar results than the state-of-the-art methods. 

Notably, our work delves into the potential usage of more novel modalities derived from the skeleton data based on the well-established CrosSCLR baseline while proposing a new teacher-student method to achieve the efficient usage of more modalities. Our method achieves superior performance improvements compared with the CrosSCLR baseline. Our approach still has the potential to be used in other state-of-the-art methods, which is recognized as future work.

\section{Conclusion}
\label{sec:conclusions}
In this paper, we present a module to improve the knowledge exchange mechanisms, which aids the exchange of implicit knowledge as distinguished from inter-sample relationships while enhancing the multi-modality model's tolerance to erroneous knowledge.
Moreover, we introduce three additional modalities derived from skeleton data and propose a novel multi-modality teacher-student model. The teacher-student model distills knowledge from a multi-modality teacher model to a student model with fewer modalities and does not lead to significant performance degradation. Experimental results of the student model have illustrated the potential for efficient use of multi-modality data based on skeleton data,
without having to regulate the performance of the added modalities and the efficiency of the entire model.
We believe our contributions can help state-of-the-art methods exploit the potential of skeleton-based multi-modality data.

\vfill\pagebreak

% -------------------------------------------------------------------------
\clearpage
\bibliographystyle{IEEEbib}
\bibliography{strings,refs}

\end{document}

%% file: tables.tex
\begin{subtable}[ht]{0.66\columnwidth}
\begin{subtable}[ht]{\columnwidth}
%\captionsetup{font={scriptsize}}
\caption{Results when different modules work.}
\vspace{-6pt}
\label{tab: IKEMworkwithEKEM}
\renewcommand{\arraystretch}{1}
\setlength{\tabcolsep}{1mm}
\resizebox{\textwidth}{!}{
\begin{tabular}{l|c}
\hline\hline 
Knowledge Exchange Module &Cross-view  \\ \hline\hline
EKEM           & 83.4             \\
IKEM           & 82.3             \\
EKEM and IKEM  & \textbf{84.1}    \\ \hline\hline        
\end{tabular}
}
\end{subtable}
\vspace{2mm}
%\hspace{2pt}

\begin{subtable}[ht]{\columnwidth}
\begin{center}
%\captionsetup{font={scriptsize}}
\caption{Results of our three-modality model, six-modality model, and baseline 3s-CrosSCLR.}
\vspace{-6pt}
\label{tab: our3sand6s}
\renewcommand{\arraystretch}{1.18}
\setlength{\tabcolsep}{1mm}
\resizebox{\textwidth}{!}{
\begin{tabular}{l|c|c}
\hline\hline
Method  & Cross-subject  & Cross-view  \\ \hline\hline
3s-CrosSCLR \cite{li20213d}               & 77.8                                                       & 83.4                                                       \\
3s-Ours & 78.0                                                       & 84.1                                                       \\ 
6s-Ours & \textbf{81.2}                                              & \textbf{86.9}                                               \\ \hline\hline
\end{tabular}
}
\end{center}
\end{subtable}
\end{subtable}
%\vspace{5mm}
%\hspace{5pt}
%\hspace{2pt}
\begin{subtable}[ht]{0.66\columnwidth}
\begin{center}
%\captionsetup{font={scriptsize}}
\caption{Results of the individual modalities of our three-modality model and six-modality model. }
\vspace{-6pt}
\label{tab: our3sand6s single modality}
\renewcommand{\arraystretch}{1.45}
\setlength{\tabcolsep}{2pt}
\resizebox{\textwidth}{!}{
\begin{tabular}{l|c|cc}
\hline\hline
Method  & Modality         &  Cross-subject  & Cross-view  \\ \hline\hline
3s-Ours & \multirow{2}{*}{joint}           & 73.3                                                       & 79.7                                                       \\ 
6s-Ours &                  & \textbf{75.5}                                              & \textbf{81.8}                                              \\ \hline
3s-Ours & motion           & 74.0                                                       & 79.5                                                       \\ 
6s-Ours & (velocity)       & \textbf{75.8}                                              & \textbf{81.1}                                               \\ \hline
3s-Ours & \multirow{2}{*}{bone}             & 73.7                                                       & 78.1                                                        \\ 
6s-Ours &                  & \textbf{74.4}                                              & \textbf{79.9}                                               \\ \hline\hline
\end{tabular}
}
\end{center}
\end{subtable}
% 
%\hspace{2pt}
\begin{subtable}[ht]{0.66\columnwidth}
\begin{center}
%\captionsetup{font={scriptsize}}
\caption{Results of our student model
and other models with the same amount of parameters in linear evaluation protocols.}
\vspace{-6pt}
\label{tab: studentandother}
\renewcommand{\arraystretch}{1.25}
\setlength{\tabcolsep}{2pt}
\resizebox{\textwidth}{!}{
\begin{tabular}{l|cc}
\hline\hline
Method          & Cross-subject & Cross-view  \\ \hline\hline
3s-CrosSCLR \cite{li20213d}       & 77.8                                                       & 83.4                                                        \\ 
3s-AimCLR \cite{guo2021contrastive}      & 78.9                                                       & 83.8                                                        \\ 
3s-HYSP \cite{franco2023hyperbolic}        & 79.1                                                       & 85.2                                                        \\ 
3s-PSTL \cite{zhou2023selfsupervised}         & 79.1                                                       & 83.8                                                       \\ 
3s-ActCLR \cite{lin2023actionlet}          &84.3                                                        &88.8                                                      \\
3s-CPM \cite{zhang2022contrastive}         &83.2                                                        &87.0                                                       \\
3s-Student-Ours & 80.7                                                       & 86.0                                     \\ \hline\hline
\end{tabular}
}
\end{center}
\end{subtable}
%\hspace{5pt}